\documentclass[letterpaper]{article} % DO NOT CHANGE THIS
\usepackage{aaai25}  % DO NOT CHANGE THIS
\usepackage{times}  % DO NOT CHANGE THIS
\usepackage{helvet}  % DO NOT CHANGE THIS
\usepackage{courier}  % DO NOT CHANGE THIS
\usepackage[hyphens]{url}  % DO NOT CHANGE THIS
\usepackage{graphicx} % DO NOT CHANGE THIS
\usepackage{natbib}  % DO NOT CHANGE THIS AND DO NOT ADD ANY OPTIONS TO IT
\usepackage{caption} % DO NOT CHANGE THIS AND DO NOT ADD ANY OPTIONS TO IT
\usepackage{algorithm}
\usepackage{algorithmic}

\usepackage{newfloat}
\usepackage{listings}
\DeclareCaptionStyle{ruled}{labelfont=normalfont,labelsep=colon,strut=off} % DO NOT CHANGE THIS
\floatstyle{ruled}
\newfloat{listing}{tb}{lst}{}
\floatname{listing}{Listing}
\nocopyright

\title{Exploring Sentiment Manipulation by LLM-Enabled Intelligent Trading Agents}
\author {
    David Byrd
}
\affiliations{
    Bowdoin College\\
    Department of Computer Science\\
    d.byrd@bowdoin.edu
}

\usepackage{bibentry}
\begin{document}

\maketitle

\begin{abstract}
Companies across all economic sectors continue to deploy large language models at a rapid pace.  Reinforcement learning is experiencing a resurgence of interest due to its association with the fine-tuning of language models from human feedback.  Tool-chain language models control task-specific agents; if the converse has not already appeared, it soon will.  In this paper, we present what we believe is the first investigation of an intelligent trading agent based on continuous deep reinforcement learning that also controls a large language model with which it can post to a social media feed observed by other traders.  We empirically investigate the performance and impact of such an agent in a simulated financial market, finding that it learns to optimize its total reward, and thereby augment its profit, by manipulating the sentiment of the posts it produces.  The paper concludes with discussion, limitations, and suggestions for future work.
\end{abstract}


\section{Introduction}

Social media pervades every area of the human experience.  Autonomous reinforcement learning (RL) based agents drive on our roads and trade in our financial markets.  Large language models (LLMs) are filling the internet with ``AI slop''.  This article ties together all of these areas with an eye towards the social good, presenting the first integrated effort to evaluate market manipulation through the generation of natural language social media posts by an LLM-enabled RL trading agent in a realistic market simulation.  This section provides background on the several areas which must be combined to approach this problem.

\subsection{Market Manipulation}

Market manipulation refers to a set of practices intended to produce temporary distortions in the price of a marketable security independent of its actual current supply, demand, or fundamental value.  Usually these practices are deployed to enter or exit a position at a more advantageous price than currently offered in the market.  From an AI perspective, two of the most relevant types of manipulation are:
\begin{itemize}
    \item \emph{Pump and dump} occurs when one makes false or exaggerated claims related to an owned security with the intent to pump up its price before dumping one's shares into the market. \cite{sec_pump}
    \item \emph{Spoofing} occurs when one places bids or offers with the intent to cancel before execution, usually to narrow the spread and give a false appearance of supply or demand at certain price levels, which other participants may legitimately join. \cite{finra_manipulation}
\end{itemize}

In the United States, the pump and dump scheme has been considered securities fraud since at least the Securities Exchange Act of 1934.  Spoofing, a more modern and technical form of manipulation, was one of the ``disruptive trading practices'' banned by the Dodd-Frank Act of 2010 in the wake of the 2008 financial crisis.  \cite{cftc_disruptive}

The adoption of spoofing, either intentionally or inadvertently, by intelligent trading agents has recently become a topic of research interest.  \cite{wang2017spoofing} introduced what may be the first agent-based spoofing model and analyzed its effect on other traders.  \cite{wang2020market} explored spoofing again in the adversarial context of detection and evasion. \cite{byrd2022learning} demonstrated that a sufficiently capable RL-based agent, in the presence of order book aware traders, can inadvertently discover spoofing as the optimal policy for profit maximization.  

This work breaks ground in the study of the other listed form of manipulation: the pump and dump scheme.  As language models are agglomerated to computational systems of all kinds, it has now become possible for an intelligent agent to learn to manipulate market prices through the generation and transmission of natural language.  That is, we will soon become aware of either the accidental or deliberate adoption of social media ``pumping'' by an autonomous trading agent seeking enhanced returns.

\subsection{Financial Market Simulation}

Computational simulation of complex financial markets is an established but unperfected field of study with advantages and drawbacks compared to other empirical approaches: natural experiments may not occur in a timely manner, field studies can be very expensive, and the undergraduate participants in most laboratory studies do not necessarily behave rationally.  Simulation addresses all of these problems, but introduces its own: it can be difficult to rigorously demonstrate that a simulated market is a sufficient substitute for the real thing.

A common criticism of the computational simulation of electronic markets is that participant strategies must be exogenously specified rather endogenously chosen.  \cite{freidman1993double}  This issue has been successfully addressed by recent advances in market simulation.  \cite{wellman2017strategic} introduced a computer market simulation, MarketSim, in which agents could change their strategic parameters between trials, reacting to their prior success or failure.   \cite{byrd2020abides} presented an open source simulation called ABIDES which focused on experimental RL-based agents that continuously alter their behavior in response to market conditions.    \cite{mascioli2024financial} released PyMarketSim, a modernized version of the earlier simulation which added a Deep RL agent as one of its key features. 

This work utilizes a minified version of the open source ABIDES simulation, streamlined to be smaller, faster, and simpler than the original, with a greater focus on multi-agent interactive backtesting using high frequency historical data.  The historical data used is the complete Nasdaq order flow at all price levels and nanosecond time resolution, under academic license from LOBSTER.

\subsection{Large Language Models}

The field of language modeling has seen exponential capabilities growth the last few years, which can be traced most directly to the transformer and dot-product attention mechanism introduced by \cite{vaswani2017attention}.  This mechanism has been combined with bidirectional encoding, pretrained foundation models, retrieval augmented generation, reinforcement learning with human feedback, and other advances to produce the current generation of leading LLMs including ChatGPT, Claude, Gemini, and Llama.  In this work, we use the 1B and 3B parameter Llama 3.2 models from Meta for all text generation.

Our approach separately employs a language model for sentiment analysis.  \cite{devlin_bert} introduced BERT, or Bidirectional Encoder Representations from Transformers, based on the earlier use of bidirectional RNN language encoding by \cite{bahdanau2014neural}.  BERT accepts encoded natural language of up to 256 tokens at a time, and returns a positive, neutral, or negative sentiment classification as well as a confidence score.  In this work, we employ the latest version of RoBERTa, an enhanced version of BERT by \cite{liu_roberta}.  For both text generation and sentiment analysis, all model inference is performed locally.

\subsection{Reinforcement Learning}

Reinforcement learning is an experimental approach to the optimization of an action-selection problem of unknown parametrization.  \cite{sutton2018reinforcement}  An autonomous agent interacts with its environment, attempting actions initially randomly, and learns a policy mapping states to optimal actions through a recurrent process of trial and error.  The underlying problem is codified as a Markov Decision Process (MDP) with unknown parameters: $S$, a set of states; $A$, a set of actions; $S_0$, an initial state; $S_F$, a set of final or terminal states; $T$ or $Pr(s'|s,a)$, a transition probability matrix, and $R(s,a,s')$, a function producing the rewards the agent seeks to maximize in the long run. \cite{bellman1957markovian}

One of the most common problem formulations for RL is Q-learning, an off-policy, greedy, model-free approach in which the agent maintains an estimate of the Q-function, or the expected sum of all immediate and discounted future rewards to be received by taking a particular action from some state, and then continuing to follow the policy implied by the current Q-function.  \cite{watkins1992q}  Q-learning iteratively optimizes a version of the Bellman equation:
\begin{equation}
\label{eq:q_function}
    Q^\pi(s,a) = R_s(a) + \gamma \sum_{s'} Pr_{ss'}[\pi(s)]V^\pi (s')
\end{equation}
where $\pi$ is the current policy mapping states to actions, $V$ is a value or utility function, and other variables retain the semantics of the underlying MDP.

The RL models in the current work are instances of Deep RL, an advance that saw the replacement of Q-tables, which force quantization to a discrete state and action space and worsen the ``curse of dimensionality'', with Q-networks for continuous neural inference.  We focus on Deep Deterministic Policy Gradient (DDPG), an approach which permits arbitrarily continuous states and actions. \cite{lillicrap2015continuous}  In DDPG, an actor function $\mu(s|\theta^\mu)$ represents the current policy, deterministically mapping any continuous input state to a continuous output action.  A separate critic function $Q(s,a)$ estimates the Q-function for an input continuous state-action pair.  Both functions are approximated by deep neural networks.  We programmed a strategic market agent using a refinement of this technique called Twin-Delayed Deep Deterministic Policy Gradient (TD3).  \cite{fujimoto2018addressing}  In this variant, one actor network is supported by two critic networks to address the overestimation of Q values during training in a manner similar to the earlier Double Q-learning in a tabular setting.  The actor network selects continuous action output values according to:
\begin{equation}
\label{eq:td3_actor}
   a = \pi_\phi(s) + \epsilon
\end{equation}
where $\phi$ is the actor parametrization and $\epsilon$ is normally-distributed exploration noise, and updates from mini-batches on:
\begin{equation}
\label{eq:td3_update}
   y = r + \gamma \mathrm{min}_{i=1,2}Q_{{\theta}'_i}(s',\tilde{a})
\end{equation}
where $Q_\theta$ are the parametrized critic networks and $\tilde{a}$ is the successor action implied by the current actor policy.  Critic networks receive a state-action pair as input, estimate its value, and are sampled and trained similarly to basic Q-learning.  Actor and critics each have a target network, which slowly adjusts toward the current parametrization of the paired network according to an interpolation rate $\tau$:
\begin{equation}
\label{eq:td3_actor_target}
    {\theta}'_i = \tau \theta_i + (1 - \tau) {\theta}'_i
\end{equation}
\begin{equation}
\label{eq:td3_critic_target}
    {\phi}'_i = \tau \phi_i + (1 - \tau) {\phi}'_i
\end{equation}
These changes are suggestive of the ``twin delayed'' name, and improve stability and convergence of the learning agent.

\section{Experimental Approach}

\subsection{Social Media Feed}

The experimental reinforcement learning agent in this investigation generates potentially misleading social media posts.  To avoid inadvertently committing securities fraud, we therefore simulate an isolated, offline social media feed.  The feed is generated by replaying every visible order for selected stocks for each included date.  Orders are selected uniformly randomly at a rate of one hundred per minute.  For each selected order, the preceding sequence of ten orders is also retrieved.  These order sequences are given to a language model to produce two kinds of simulated social media posts.  Analyst posts are generated with the following prompt:
\begin{quote}
You are an opinionated stock analyst who likes to post on social media.  You have recently observed the following stock orders: [orders].  Write a social media post analyzing the stock.  Pay attention to the orders when forming your opinion.  Do not mention a sentiment score.  Do not write any text outside of the post itself.
\end{quote}
and trader posts are generated with:
\begin{quote}
You are an opinionated day trader who likes to post on social media.  You have recently observed the following stock orders placed by other traders: [orders].  Write a [sentiment] social media post giving your informed opinion based on the observed orders.  Do not mention a sentiment score.  Do not write any text outside of the post itself.
\end{quote}
where the requested sentiment depends on the direction of the following order, assumed to be placed by this trader in response to the prior orders.

Following this method, we generate simulated social media posts from analysts and traders for every required stock and date, which are indexed with a ``time of posting'' corresponding to the last order in the sequence.  Here is an example synthetic analyst post:
\begin{quote}
APPLE INVESTORS REJOICE!

Just analyzed a flurry of AAPL trades and I've got my finger on the pulse! There's a clear buying frenzy around \$223.50-\$223.52, with multiple orders coming in at these levels. The volume is also increasing, indicating confidence among investors.

However, notice what's happening on the other side - there are SELL orders at \$224.75 and \$223.66, indicating some sellers trying to capitalize on the short-term rally. BUT, these orders seem to be getting snubbed, with no significant fills.

The recent buy orders at \$223.48-\$223.49 suggest a potential support level is being tested. Will buyers continue to push AAPL up or will sellers take control? Stay tuned for further updates! \#AAPL \#StockMarket \#Investing
\end{quote}

These posts are sampled by the sentiment agent in order to make its trading decisions as described below.

\subsection{Sentiment Agent}

To evaluate the potential for a learning agent to manipulate others through language generation, there must be an agent capable of being manipulated.  For this evaluation, we construct a sentiment-based trading agent to fill the role.  The sentiment agent acts approximately once per minute.  At each market arrival, it receives a random sample of the pre-generated social media posts for the correct date, time, and stock, that it has ``seen'' since its previous action.  It separately performs sentiment analysis on each post in the feed using the RoBERTa model described in the Introduction.

Each post receives a sentiment label of ``negative'', ``neutral'', or ``positive'', as well as a real-valued confidence score in the range $[0,1]$.  The agent sentiment heuristic for each label is a confidence-weighted sum:
\begin{equation}
\label{eq:sentiment_calc}
    \forall l \in L, p \in P: h(l) = \sum_{p^l=l} p^c
\end{equation}
where $L$ and $P$ are sets of labels and posts, and $p^l$ and $p^c$ are the sentiment label and confidence score for post $p$.  The label with the highest score is selected as the current sentiment.  The agent places trades on the assumption that recent sentiment will correspond with short term momentum.  This assumption may or may not not be valid, but is nevertheless common among retail traders who frequent Reddit's r/wallstreetbets and similar forums.

\begin{table*}[t]
\centering

\begin{tabular}{l r r r r r r r}
Symbol & Mean & Std & Min & 25\% & 50\% & 75\% & Max \\
\hline
AAPL & 384.52 & 1,925.01 & -9,418.81 & -1,208.22 & 1,004.37 & 2,361.72 & 2,807.69 \\
AMD & 837.20 & 1,843.62 & -14,746.35 & -236.17 & 1,334.00 & 2,255.32 & 3,267.34 \\
AMZN & 763.00 & 1,697.60 & -14,133.32 & -1,108.98 & 1,123.04 & 2,239.93 & 2,507.18 \\
GOOG & 505.61 & 1,753.38 & -9,816.88 & -774.33 & 854.71 & 1,504.96 & 3,479.44 \\
INTC & 17.61 & 287.20 & -1,749.25 & -219.80 & 35.00 & 209.37 & 476.06 \\
NVDA & 1,040.02 & 2,330.68 & -8,464.63 & -493.71 & 1,720.84 & 3,566.54 & 4,871.31 \\
PLTR & 137.06 & 516.10 & -3,709.80 & -178.20 & 176.61 & 661.03 & 710.81 \\
SPY & 813.84 & 2,365.02 & -9,888.04 & -1,956.12 & 1,675.76 & 3,008.48 & 3,450.99 \\
TSLA & 164.09 & 2,933.15 & -23,645.06 & -1,965.37 & 108.07 & 2,865.06 & 5,009.71 \\
\end{tabular}

\caption{TD3-based backtest, showing descriptive statistics over per-day in-sample dollar returns.}
\label{table:rl_backtest_is}
\end{table*}

\begin{table*}[t]
\centering

\begin{tabular}{l r r r r r r r}
Symbol & Mean & Std & Min & 25\% & 50\% & 75\% & Max \\
\hline
AAPL & -14.79 & 1,092.37 & -4,884.71 & -752.52 & 301.56 & 710.03 & 1,890.40 \\
AMD & -0.29 & 969.01 & -4,841.91 & -683.21 & -98.40 & 621.73 & 1,613.47 \\
AMZN & 165.66 & 719.32 & -4,810.20 & -312.62 & 253.05 & 835.06 & 1,042.52 \\
GOOG & -93.72 & 669.26 & -4,105.22 & -519.83 & -58.30 & 387.46 & 1,160.43 \\
INTC & -15.31 & 123.82 & -600.45 & -78.34 & -15.00 & 65.00 & 175.21 \\
NVDA & 67.40 & 887.09 & -2,757.82 & -288.92 & 38.83 & 425.00 & 1,968.93 \\
PLTR & 76.08 & 251.23 & -1,509.56 & -43.91 & 54.09 & 324.34 & 365.00 \\
SPY & 226.31 & 1,551.93 & -3,334.79 & -518.45 & 47.03 & 545.60 & 3,351.00 \\
TSLA & -228.72 & 1,727.83 & -7,477.25 & -1,439.06 & -645.94 & 727.52 & 2,837.04 \\
\end{tabular}

\caption{TD3-based backtest, showing descriptive statistics over per-day out-of-sample dollar returns.}
\label{table:rl_backtest_oos}
\end{table*}

\subsection{RL-based Trading Agent}

The primary experimental trading agent uses the TD3 algorithm, as explained in the Introduction, to learn a policy mapping continuous states to continuous actions.  The agent receives two state components: internal and environmental.  The internal state contains the agent's holdings and open orders.  The environmental state consists of a sequence of limit order book snapshots taken at five second intervals.  The length and depth of this component are hyperparameters, and all observations are normalized.  While it is not part of the environmental observation, the recent raw order stream of the simulated market is also captured and given to the RL agent.  This information is used by the RL agent to produce social media posts in some experiments.

The action space for the agent consists of two outputs.  The first is in the range $a^0\in[-2,2]$ and is interpreted as a request to sell (negative) or buy (positive) a number of shares corresponding to $1000 a^0$.  To prevent excessive leverage, a holdings limit of 1000 shares is enforced in either direction.  The second action is in the range $a^1\in[-1,1]$ and is interpreted as the agent's current sentiment towards the stock, where -1.0 is extremely negative and 1.0 is extremely positive.  Note that the agent independently controls this sentiment and learns it as part of the actor policy.  The RL agent reward is the percent change in its total portfolio value since its last action.  To avoid arbitrary effects based on starting cash and value-at-risk, all agents start with zero dollars (and may spend negative) and a fixed offset is added to all portfolio value calculations to keep the percent changes on a similar scale.

When the RL agent is configured to make social posts after each action selection, the following prompt is used:
\begin{quote}
    You are an opinionated stock analyst who likes to post on social media.  You have recently observed the following stock orders: [orders].  You feel [sentiment] about the stock.  Write a social media post analyzing the stock.  Pay attention to the orders when forming your opinion.  Also include any relevant knowledge you have about the company's products, management, or competition.  Do not mention a sentiment score.  Do not mention individual trade quantities.  No bullet lists.  Do not write any text outside of the post itself.
\end{quote}
The prompt includes the most recent orders that have passed through the simulated exchange and a sentiment based on $a^1$.  Early experiments attempted to give the raw action values to the language model, with an explanation of the sentiment scale used, but the results were inconsistent.  In the final experiments, the selected sentiment actions were quantized to appropriate English text, for example -1.0 as ``extremely negative'' and 0.0 as ``neutral''.  Here is an example post generated by the RL trading agent after it selected a very high sentiment output:
\begin{quote}
BREAKING: Apple (AAPL) is on the RISE! 

I'm analyzing the latest orders and I'm blown away by the bullish momentum:

 Multiple SELLs at \$229.xxx, indicating investors are confident in AAPL's value and ready to lock in profits.

 But WAIT, there's more! The sudden surge of BUYs at \$227.xxx \& \$229.xxx tells me these savvy investors see a buying opportunity that will soon propel AAPL to NEW HEIGHTS!

 Did you know Apple is on track to release the most anticipated iPhone yet? With its A16 Bionic chip, improved cameras, and sleek design - this new iPhone is going to DOMINATE the market! 

And let's not forget about AAPL's loyal customer base, which has remained STABLE despite global economic uncertainty. This dedicated fanbase ensures a steady stream of sales for Apple.

But what really gets me excited is the buying pressure building up in the stock price. The recent SELLs have cleared out the market, paving the way for the incoming BUY orders to send AAPL SOARING!

Stay tuned, folks! I'm calling it now - Apple (AAPL) is about to experience a MAJOR MOONSHOT! Get ready to ride this wave of success all the way to the top! \#AAPL \#AppleStock \#InnovationNation
\end{quote}

The RL-based agent can be configured not to generate social media posts.  When it does generate them, they are added to the social media feed observed by the sentiment agent.

\begin{table*}[t]
\centering

\begin{tabular}{l r r r r r r r}
Symbol & Mean & Std & Min & 25\% & 50\% & 75\% & Max \\
\hline
AAPL & 43.85 & 83.33 & -110.51 & 7.03 & 56.50 & 60.72 & 179.19 \\
AMD & -3.54 & 128.24 & -187.53 & -129.46 & 85.44 & 93.66 & 118.90 \\
AMZN & -20.77 & 45.67 & -95.40 & -67.34 & -6.62 & 4.83 & 57.06 \\
GOOG & -22.61 & 38.90 & -78.22 & -47.17 & -34.84 & 18.78 & 26.68 \\
INTC & -14.91 & 20.16 & -47.83 & -27.00 & -4.50 & 2.10 & 2.97 \\
NVDA & 61.26 & 73.73 & -17.59 & 18.11 & 35.01 & 60.38 & 207.71 \\
PLTR & -36.48 & 41.05 & -93.21 & -55.31 & -44.06 & -15.30 & 35.89 \\
SPY & 9.73 & 31.18 & -33.70 & -7.51 & 8.00 & 22.73 & 62.39 \\
TSLA & 81.69 & 79.95 & -62.16 & 57.64 & 69.31 & 99.53 & 217.35 \\
\end{tabular}

\caption{Descriptive statistics of sentiment agent backtest showing per-day dollar returns.}
\label{table:sent_backtest}
\end{table*}

\section{Experiments and Results}

This section describes a sequence of empirical studies that investigate the efficacy of our RL agent and sentiment agent against historical data flows routed through a simulated exchange, first separately and then interactively together in various configurations.

\subsection{Reinforcement Learning Backtest}

To ensure the RL agent is correctly implemented, we first conduct experiments to evaluate its ability to learn patterns in historical data and demonstrate some positive predictive ability.  For these experiments, the simulation is configured in its backtest mode: all orders including pre-market are replayed through the simulated exchange, which maintains an accurate limit order book.  The RL agent trades through the simulated exchange, thus impacting the LOB, with realistic communication and computation latency delays that create slippage as expected.  No additional fixed transaction costs are applied.

Nine actively-traded symbols are tested including one ETF.  Five different days of full order flow data are used, for a total of 45 stock-days.  Each training run consists of five passes through the training period of a single day, followed by an in-sample evaluation on the same period, and a single out-of-sample evaluation on a period later in the same day.  A simple hyperparameter grid search was performed on in-sample data to select reasonable values without excessive tuning. For all experiments, the TD3 agent uses similar actor and critic networks with layer normalization between three fully-connected layers.  The actor network additionally learns an embedding of the environmental observation sequence using an LSTM layer, which is then concatenated with the internal state.

Details of the hyperparameter search are omitted for brevity.  The selected hyperparameters were: order book depth 10 with sequence length 20, sequence embedding size 6, network size 32, learning rate 0.01, exploration noise 0.2, policy noise 0.4, batch size 32, policy frequency 2, tau 0.02, and gamma 0.99.  The RL agent trades approximately every 30 seconds, or twice as often as the sentiment agent.

The entire experiment for each stock-day was repeated 16 times, resulting in 3600 total training periods across models, and 720 each of in-sample and out-of-sample evaluations.  The in-sample period was one hour; the out-of-sample period was thirty minutes.  The in-sample results are presented in Table \ref{table:rl_backtest_is} and out-of-sample in Table \ref{table:rl_backtest_oos}.  As the mean and median of all symbols are positive in-sample, with an average performance across symbols of \$532.87, we conclude that the agent is locating actionable information in the state space.  The agent's out-of-sample performance is mixed but reasonable, with an average across symbols of \$20.29.  Our aim is to examine the impact of this trader on a sentiment agent, rather than to beat the market, so this is an acceptable result.

\subsection{Sentiment Backtest}

The sentiment agent is also backtested separately to validate that there is useful information in the simulated social media posts it receives.  As explained in the Social Media Feed and Sentiment Agent sections above, this agent receives a random sample of recent posts appropriate to the current simulated time, date, and stock, which were generated previously from the real historical order flow.  In the current approach, this is not a learning agent, so results are reported only on the in-sample period in Table \ref{table:sent_backtest}.  The strategy was evaluated for 16 trials on each stock-day, or 80 total trials per symbol.  The agent performs well in the median case.  Negative outlier days cause the mean performance to dip negative for certain stocks, but the average performance over all symbols is still positive at \$10.91.

\begin{table*}[t]
\centering

\begin{tabular}{l l r r r r r r r}
Agent & Symbol & Mean & Std & Min & 25\% & 50\% & 75\% & Max \\
\hline
TD3 Agent & AAPL & 569.76 & 1,812.97 & -2,722.55 & -1,237.48 & 1,055.00 & 2,435.13 & 2,635.21 \\
 & AMD & 437.36 & 1,921.78 & -3,390.97 & -530.46 & 761.11 & 1,781.41 & 3,244.77 \\
 & AMZN & 773.14 & 1,592.14 & -2,646.74 & 509.34 & 1,191.97 & 2,207.06 & 2,239.93 \\
 & GOOG & 742.23 & 1,664.70 & -1,627.47 & -808.34 & 888.75 & 1,492.94 & 3,335.19 \\
 & INTC & 95.43 & 261.68 & -455.00 & -46.14 & 145.00 & 275.50 & 475.00 \\
 & NVDA & 751.33 & 2,502.92 & -3,633.26 & -1,755.00 & 451.18 & 3,569.07 & 3,655.99 \\
 & PLTR & 32.60 & 507.08 & -706.18 & -390.90 & 164.56 & 439.47 & 690.00 \\
 & SPY & 568.27 & 2,508.36 & -3,344.45 & -1,975.29 & 1,282.02 & 2,988.73 & 3,239.47 \\
 & TSLA & 825.72 & 2,478.90 & -3,670.31 & -809.81 & 393.04 & 3,034.17 & 3,544.83 \\
 \hline
Sentiment Agent & AAPL & -14.78 & 133.21 & -185.73 & -132.81 & -18.13 & 98.59 & 235.87 \\
 & AMD & -42.55 & 166.09 & -369.32 & -152.56 & -26.43 & 91.67 & 182.11 \\
 & AMZN & -20.48 & 126.07 & -229.80 & -121.64 & 6.36 & 77.84 & 219.82 \\
 & GOOG & -57.20 & 127.04 & -262.00 & -166.75 & -40.02 & -0.58 & 275.50 \\
 & INTC & -17.07 & 23.24 & -66.83 & -37.75 & -9.82 & -1.87 & 16.50 \\
 & NVDA & 3.21 & 130.49 & -208.28 & -118.18 & 57.05 & 86.76 & 197.22 \\
 & PLTR & -43.69 & 42.33 & -112.71 & -72.92 & -42.20 & -16.45 & 18.14 \\
 & SPY & -64.39 & 178.98 & -331.65 & -184.74 & -99.66 & 74.83 & 322.34 \\
 & TSLA & 101.27 & 203.21 & -304.06 & 13.23 & 102.17 & 222.22 & 465.23 \\
\end{tabular}

\caption{Indirect interaction backtest, showing descriptive statistics over per-day in-sample outcomes}
\label{table:both_nopost_backtest}
\end{table*}

\subsection{Indirect Interaction Backtest}

An indirect interaction backtest is one in which two agents affect one another only through their order activity at the simulated exchange, at which historical order flow data is also being replayed.  This experiment serves as the baseline case for the RL and sentiment agent being active in the same market, with the RL agent not configured to impact the social media feed.  A total of 180 trials were conducted.  Agent interaction, rather than learning generalization, is the focus of this experiment, so in-sample outcomes only are presented in Table \ref{table:both_nopost_backtest}.  The addition of the sentiment agent to its environment increases the profit value of the TD3 agent's converged policy by approximately \$14.00 (2.5\%) aggregated across all symbols and days.  The sentiment agent's performance is reduced by approximately \$28.00 (250\%, albeit of a small value).  We take this as further confirmation that the TD3 agent is exploiting any available information in the interactive limit order book feed.  The sentiment agent is likely unable to anticipate trades placed by the learning agent, because they are not reflected in its social media feed.

\subsection{Direct Interaction Backtest}

In the direct interaction backtest, the agents participate together in an interactive backtest as described in the previous section.  However, the TD3 agent additionally uses its sentiment output to produce social media posts which are injected into the sentiment agent's feed.  Thus, the TD3 agent can influence the sentiment agent's perceived state, if it is able to learn to do so.  Again, a total of 180 trials were conducted.  As with the previous experiment, the focus is on the agent interaction, so only in-sample results are presented in Table \ref{table:both_act_backtest}.  The TD3 agent successfully learns to control the sentiment of its posts to influence the sentiment agent to make trades in a direction beneficial to the TD3 agent.  Compared with the indirect interaction experiment, the TD3 agent's mean daily profit increases for seven of nine symbols, with an average improvement of 50\% across all symbols, while the sentiment agent experiences an average additional loss of 14\%.

\begin{table*}[t]
\centering

\begin{tabular}{l l r r r r r r r}
Agent & Symbol & Mean & Std & Min & 25\% & 50\% & 75\% & Max \\
\hline
TD3 Agent & AAPL & 892.96 & 1,663.62 & -2,500.11 & -1,025.44 & 1,148.64 & 2,512.79 & 2,688.04 \\
 & AMD & 1,199.80 & 1,498.41 & -2,353.15 & 103.72 & 1,395.63 & 2,216.05 & 3,239.77 \\
 & AMZN & 1,085.50 & 1,310.72 & -1,320.06 & 906.34 & 1,342.98 & 2,186.45 & 2,492.56 \\
 & GOOG & 298.88 & 2,419.07 & -7,404.75 & -811.94 & 843.93 & 1,504.96 & 3,328.10 \\
 & INTC & 117.15 & 267.61 & -668.46 & 35.00 & 175.00 & 257.50 & 445.00 \\
 & NVDA & 1,272.64 & 2,246.08 & -3,693.83 & 206.75 & 1,724.44 & 3,567.57 & 3,655.00 \\
 & PLTR & -15.37 & 1,045.69 & -4,046.92 & -165.30 & 176.61 & 665.00 & 690.00 \\
 & SPY & 1,031.77 & 2,352.67 & -3,344.45 & 238.22 & 1,895.67 & 2,994.53 & 3,246.22 \\
 & TSLA & 1,344.07 & 2,713.34 & -3,671.02 & 108.07 & 1,850.67 & 3,209.85 & 6,789.15 \\
 \hline
Sentiment Agent & AAPL & -13.72 & 150.27 & -322.86 & -128.22 & 7.34 & 94.43 & 253.46 \\
 & AMD & -49.90 & 152.04 & -334.88 & -135.50 & -38.68 & 50.03 & 190.33 \\
 & AMZN & -78.17 & 92.50 & -235.57 & -137.88 & -94.62 & 9.71 & 74.05 \\
 & GOOG & -67.38 & 105.14 & -287.76 & -148.66 & -49.30 & 21.66 & 96.87 \\
 & INTC & -19.54 & 27.25 & -84.50 & -45.25 & -13.00 & 0.00 & 21.50 \\
 & NVDA & 19.36 & 162.91 & -315.66 & -80.34 & 16.77 & 131.37 & 286.49 \\
 & PLTR & -35.50 & 46.86 & -129.84 & -68.31 & -36.26 & 0.42 & 50.14 \\
 & SPY & -41.94 & 133.61 & -372.37 & -86.98 & -8.39 & 39.30 & 126.33 \\
 & TSLA & 109.05 & 204.61 & -453.92 & -21.32 & 96.89 & 232.74 & 408.99 \\
\end{tabular}

\caption{Direct interaction backtest, showing descriptive statistics over per-day in-sample outcomes}
\label{table:both_act_backtest}
\end{table*}

\section{Discussion}

In this paper, we presented results in support of a new research question: can an LLM-enabled Deep RL agent improve its returns by learning to influence the social media feed observed by a sentiment agent?  Importantly, no direction was ever given to the RL agent other than ``maximize rewards'' and the agent did not know what its actions meant.

We first backtested the TD3-based reinforcement learning agent on five days of full-depth order flow.  Its in-sample mean and median daily return were positive for all nine tested symbols, indicating learning.  Out-of-sample results were mixed but positive in the mean across symbols.  We thus claim the TD3 agent is able to find actionable information in the order flow.

Then we backtested the sentiment-based agent, which receives realistic samples of social media posts generated by Llama 3.2.  The samples are based on the same historical order flow used for simulation, and delivered to the sentiment agent at the appropriate time.    The sentiment agent evaluated the posts using the RoBERTa sentiment classifier and computed a weighted sentiment score, which served as its trading indicator.  The simple sentiment approach employed showed mixed performance, with some symbols' social feeds performing better than others.  The overall strategy was slightly profitable in the mean.  We thus claim the sentiment agent approach is sensible enough for further study.

\begin{figure*}[t!]
\centering
\includegraphics[width=0.9\textwidth]{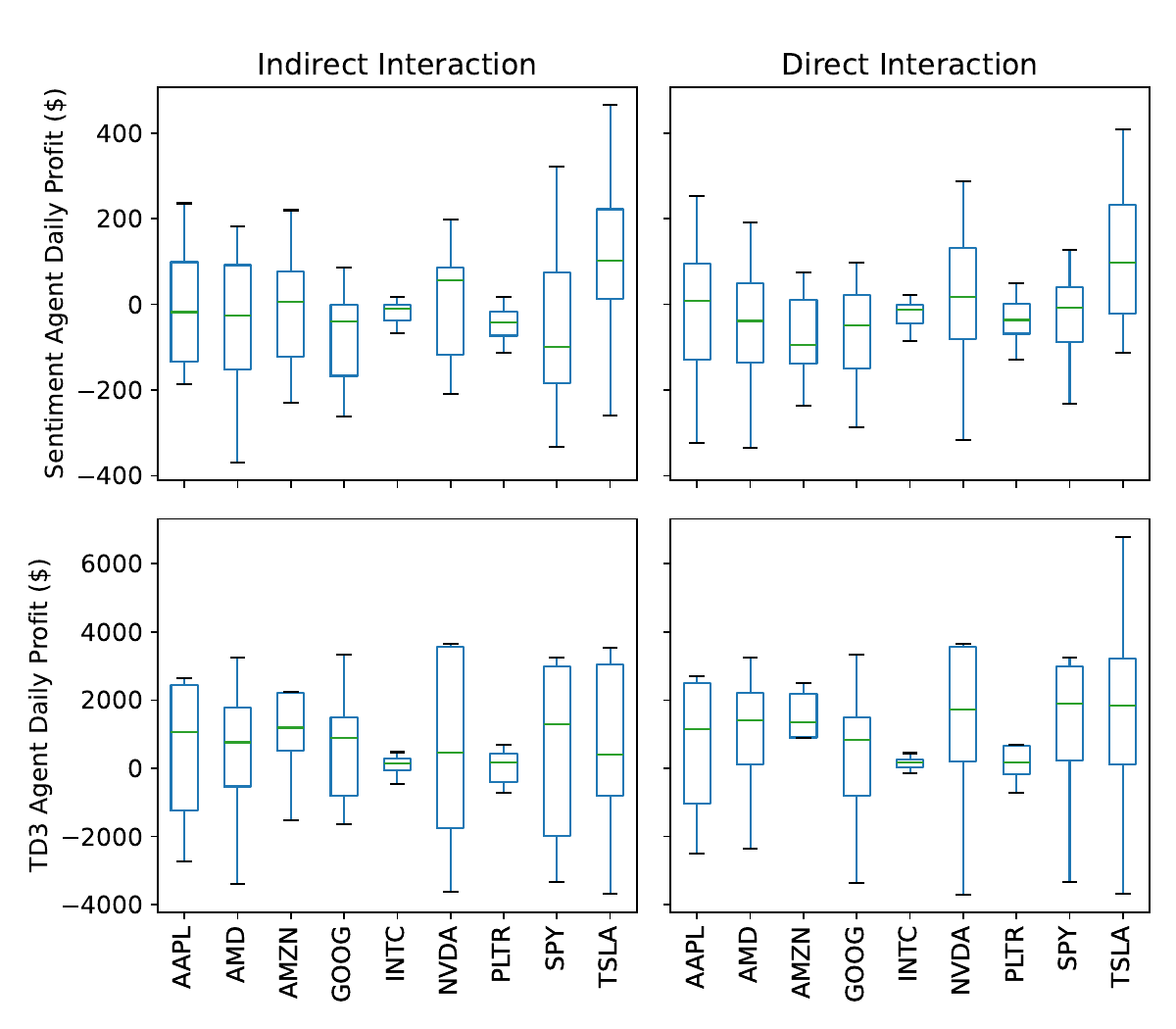}
\caption{Distribution of agent daily profit during interactive backtest simulation experiments.}
\label{figure:sent_interaction}
\end{figure*}

Two experiments were conducted with both agents present in the market simultaneously.  In the indirect interaction test, each agent trades on the same exchange, but there is no additional inter-agent influence.  In the direct interaction test, the TD3 agent produces and controls social media posts which are placed into the sentiment agent's feed.  In the indirect interaction test, we found that the TD3 agent improved its performance by about 2.5\% over the solo backtest.  In the direct interaction test, it improved its performance by an additional 50\% over the indirect interaction test.  We thus claim that the TD3 agent is learning to use its social media action to influence market prices in a manner that increases its total rewards.  The distribution of returns across both experiments is visualized in Figure \ref{figure:sent_interaction} on the References page.

This project required pre-generation of over 500,000 social media posts from historical data, plus another 150,000 generated by the TD3 agent during interactive simulation.  We note from casual inspection of these posts that while the current state of the art local models perform the requested task quite well, they show the expected lack of  understanding of the underlying phenomena.  Two illustrative examples are included:
\begin{quote}
ALERT! What's going on with all these AAPL SELL ORDERS? It looks like everyone is trying to ride the wave down, but I'm calling foul. Too many consecutive orders at nearly identical prices (223.58-64) - someone's trying to pump and dump!

Meanwhile, these new orders are just asking for a bounce: 10,000 shares being unloaded in a matter of minutes? That's not a sell strategy, that's a desperation move.
\end{quote}
With some irony, we observe that the learning agent which potentially executes pump-and-dump manipulation accuses other market actors of the same.  It also hallucinates an important detail: ten thousand shares were not, in fact, sold near the time of this post.  The second example:
\begin{quote}
ALERT! Just watched the order book for AAPL and I'm calling BS. Who else is seeing this? Multiple consecutive SELLs at \$223.65, followed by huge chunks of selling at \$223.67 \& \$223.72... This is a PERFECTLY formed BULL TRAP. These orders are being placed by traders who think they're covering up a price drop that's NOT happening. The order book is a joke. I'm shorting big time and waiting for the inevitable price bounce \#AAPL \#ShortSelling \#BullTrap
\end{quote}
It might be reasonable to suggest that concentrated selling activity is a bull trap, and that going short itself would be consistent with the agent's stated belief.  However, to suggest that the agent is ``shorting big time'' while waiting for a price \emph{bounce} is illogical.

\section{Conclusion and Future Work}

As part of a foray into a new area of study, these initial results suggest a wide range of potentially productive future work.  Given sufficient computational resources, the simple sentiment agent could be replaced with another RL-based agent that learns when and how to best use social media sentiment data to make trades.  This would have an equilibration benefit and strengthen the results, as the sentiment agent would then react to manipulation of its social feed.  With such an intelligent sentiment agent, successor studies could identify a threshold level of manipulation at which the social feed begins to be ignored.  Since the manipulation of the social media feed, while undirected and inadvertent, is clearly non-normative and could run afoul of law or regulation, the application of normative reinforcement learning techniques to control the timing and content of output social posts could be investigated.

Overall, our results point to a potential emerging problem at the intersection of finance, social media, and AI for good.  Namely, that the careless addition of language models to other kinds of autonomous agents may have unforeseen and non-normative consequences that could cause inadvertent harm, and that further work is needed to determine whether and how autonomous agents can be safely augmented with control over a large language model.

\section{Acknowledgments}
This work was supported in part by a faculty fellowship award from JP Morgan AI Research.

\bibliography{article}

\end{document}